# Adversarial Attacks, Regression, and Numerical Stability Regularization

**Andre T. Nguyen** and **Edward Raff**
{Nguyen_Andre, Raff_Edward}@bah.com
Booz Allen Hamilton

## Abstract

Adversarial attacks against neural networks in a regression setting are a critical yet understudied problem. In this work, we advance the state of the art by investigating adversarial attacks against regression networks and by formulating a more effective defense against these attacks. In particular, we take the perspective that adversarial attacks are likely caused by numerical instability in learned functions. We introduce a stability inducing, regularization based defense against adversarial attacks in the regression setting. Our new and easy to implement defense is shown to outperform prior approaches and to improve the numerical stability of learned functions.

## 1 Introduction

Adversarial machine learning has been a research area for over a decade (Biggio and Roli 2017), but has received increased attention in the past few years. In the adversarial classification case, some malicious adversary may attempt to trick or fool a classifier into producing an incorrect decision. This has natural applications and importance to domains with real-world adversaries such as spam detection (Lowd and Meek 2005) and malware (Fleshman et al. 2018), and the potential to interfere with autonomous systems such as self-driving cars (Eykholt et al. 2018).

Despite this newfound investment in adversarial machine learning, we have found that relatively few works investigate the regression case. Regression is a common goal and component of many systems, and so in this work we seek to better understand the adversarial problem from the regression perspective and provide an improved defense technique.

Given a labeled data set $\{x_i, y_i\}^N$ of $N$ values, the goal is to predict a real-valued output $y_i$ given the feature vector $x_i \in \mathbb{R}^D$. Given a neural network parameterized by $\theta$, the output of the network becomes $\hat{y}_i = f_\theta(x_i)$. To train a network, a loss $\ell(\cdot)$ is minimized. The $L_2$ or squared error loss, $\ell(y - \hat{y}) = (y - \hat{y})^2$, is a common choice.

Our goal in this work is to better understand how to defend a neural network model $f_\theta(\cdot)$ against adversarial perturbations, catalog the differences between adversarial regression and classification, and develop improved defenses. In particular, we find that approaching the problem from the perspective of *numerical stability* is informative and allows us to devise a new defense that outperforms prior approaches in this space.

The numerical (in-)stability of an algorithm broadly describes the degree to which a function's output may change with changes in the input. This is a common concern when implementing numerical libraries and even when implementing neural networks[1]. We propose the perspective that the learned network $f_\theta(\cdot)$ itself is converging to a numerically unstable solution, which then allows one to readily find adversarial examples.

We propose a regularization based defense by devising a penalty that encourages stability in learned functions. The related work to our own will be discussed in section 2, and the derivation of our technique in section 3. Our methodology for evaluation will be reviewed in section 4 and results in section 5, followed by our conclusions in section 6.

## 2 Related Work

Biggio and Roli (2017) provided an extensive survey of the history of adversarial machine learning and current research directions. Even with the recent surge in interest, surprisingly little work has been done on adversarial regression problems. Tong et al. (2018) looked at attacks in the special case of an ensemble of linear models. While informative and able to make stronger provable statements, the linear case doesn't capture the larger class of non-linear problems we are interested in. The focus on only special cases of regression is a common theme. For example, others have looked at important special case problems like selecting an optimal threshold against an adversary for regression tasks (Ghafouri, Vorobeychik, and Koutsoukos 2018). In contrast, our goal is to produce a defense that is generically useful to reduce the effectiveness of adversarial attacks.

As we take a numerical stability perspective to adversarial machine learning, we find that little work has taken the same approach directly — but many have looked at solutions from related views. Singh, Sinha, and Krishnamurthy (2018) in particular look at the hypothesis that the weights of individual layers of a network are ill-conditioned. Their work showed only minor improvement in robustness when penalizing the condition number $\kappa$ of the hidden layers. In contrast to our work, we are not concerned with the numerical stability of individual layers, but with the numerical stability of the network as a whole function.

---

[1] The log-sum-exp trick is a common method of computing the softmax loss while avoiding numerical instability.

In the classification context, it is desirable that the loss function $\ell(\cdot)$ should have no change in output for small perturbations $\Delta$ (i.e., the predicted class should remain unchanged). Lyu, Huang, and Liang (2015) phrased this goal as the min-max optimization problem in Equation 1.

$$\min_{\theta} \max_{\Delta: \|\Delta\|_p < \epsilon} \ell(f_\theta(x + \Delta)) \quad (1)$$

This approach makes intuitive sense as a goal to optimize. The maximization portion finds the perturbation $\Delta$ that maximizes the loss, and the outer minimization attempts to mitigate this loss by optimizing the parameters $\theta$. Lyu, Huang, and Liang (2015) and others (Simon-Gabriel et al. 2018; Yu et al. 2018) have attempted to optimize this objective by approximation via a truncated Taylor series, giving regularizers of the general form shown in Equation 2, where $p* = p/(p-1)$ is the dual of $p$.

$$\epsilon \cdot \left\| \frac{\partial f(x)}{\partial x} \right\|_{p^*} \quad (2)$$

This form is equivalent to the absolute condition number with respect to the $p^*$ norm. In this new perspective we can argue these prior works are directly attempting to optimize the solution with respect to a common notion of numerical stability. However, these works have been designed for the classification scenario – where we desire *no* change in response for a small change in input. For regression problems, we must expect *some* change in response for any change in the input. As such, our approach better captures this belief for the regression case. Our approach will also avoid any use of the Hessian, avoiding the need for more sophisticated techniques like double-backpropagation (Drucker and Le Cun 1991) in order to compute our regularization term.

In a different vein, more advanced optimization techniques have been explored to provide *provably* robust defenses against adversarial attacks. A number of works have recently shown this is possible and developed defenses (Abbasi and Vision 2018; Dvijotham et al. 2018; Wong and Kolter 2018; Wong, Schmidt, et al. 2018). While these defenses still have improvements to be made in terms of scalability and fully capturing potential threat models, they are a promising avenue for resolving the adversarial attack issue with regard to classifiers. However, these solutions rely on the property that a classifier's predictions should have *zero* change under small perturbations. In regression tasks, the output must necessarily change with input perturbations, and as such it is not obvious to us on how to adapt these new approaches to the regression space.

## 3 Adversarial Numerical Stability Regularization

We take the perspective that adversarial attacks are potentially a symptom of *numerical instability* in the found solution. For a numerically stable function $f(\cdot)$, we would expect that $|f(x) - f(x + \Delta)| \leq \epsilon$, where $\Delta$ is a small offset to $x$ and $\epsilon$ some threshold. Most works on adversarial attacks have looked at making small perturbations to the input that cause dramatic changes in the output, violating the numerical stability expectation.

As such, we seek to increase numerical stability by regularizing the network to encourage numerical stability. More explicitly, we seek to encode the idea that the output of two points that are near each other should be similar. We start by defining an adversarial robust loss as (3). Here we use $\|\Delta_x\|_p < \epsilon$ to denote that we are sampling a point $\Delta_x$ uniformly from the $p$ norm ball of radius $\epsilon$.

$$\ell(y - f_\theta(x)) + \lambda \cdot \mathop{\mathbb{E}}_{\Delta_x: \|\Delta_x\|_p < \epsilon} \left[ \ell\left( f_\theta(x) - f_\theta(x + \Delta_x) \right) \right] \quad (3)$$

The goal is that the loss is composed of two components: the original loss $\ell$ applied to the difference between the target output $y_i$ and the predicted output $\hat{y}_i = f_\theta(x_i)$, and the regularization penalty. Our regularization term desires that the expectation of the output between a point $x$ and all points within an $L_p$ ball with radius $\epsilon$ around $x$ are the same.

Intrinsically, if the output of the neural network is a constant this regularization term will have no penalty and no adversarial attack is possible with respect to the $L_p$ ball. But a constant function is not useful, and so the regular loss term encourages differentiation. The goal is that balancing these penalties will encourage the response of the network $f_\theta(\cdot)$ to have only the degree of variation needed. Rephrased another way, this penalty imposes a prior belief that the response $\hat{y}$ of the network should be smooth and flat — and corresponds well with our intuition that adversarial attacks are related to numerical instability. The value $\lambda \in \mathbb{R}_{\geq 0}$ then controls how strong this regularization penalty is.

We note that compared to prior works which regulate the gradient, our use of the expectation over the $L_p$ ball may seem as a weaker prior. The Gradient Regularization penalties are derived as *approximations* to a penalty with respect to the perturbation that would produce the maximal magnitude response (i.e., the most numerically unstable point). Our expectation term must cover *all* of the $L_p$ ball, and may not adequately cover the region in which the adversarial examples exist, but could provide better coverage than the approximation of the maximal point. The benefit of using our expectation though is that we now directly tackle our causal hypothesis, numerical stability, with a penalty that is easy to implement via sampling.

### 3.1 Incorporating Local Flexibility

While we have defined the general form of our regularizer in (3), we note a potential shortcoming in the indiscriminate application of the $\lambda$ term. Since we are performing a regression based problem, we know that some minimal degree of change in output is expected for any perturbation $\Delta$. Regardless of the value of $\lambda$, we will end up imposing a penalty on even small reasonable changes in the output of $f(x)$.

Prior works that used Gradient Regularization share this same shortcoming. As discussed in section 2, they can be seen as penalizing the absolute condition number of the learned function. But if the learned function $f_\theta(\cdot)$ has a condition number lower than the true function $f^*(\cdot)$, than our regularization is over-zealous and forcing under-fitting of the model.

If we had the idealized ground-truth function $f^*(\cdot)$, it would make sense that we should impose no penalty when $|f_\theta(x) - f_\theta(x+\Delta)| \leq |f^*(x) - f^*(x+\Delta)|$. Ideally the values should be equal, and when our approximation is smaller it means it has not yet fully modeled the underlying function $f^*(\cdot)$. In this case the model needs to increase its complexity, and so we do not wish to penalize it.

Since we do not have $f^*(\cdot)$, we seek to approximate the threshold by looking to our training data. If $x_i$ is the training datum under consideration, let $x_i^{(1)}$ denote it's nearest neighbor in the training data (i.e., $x_i^{(1)} = \mathrm{argmin}_{j, \forall j \neq i} \|x_j - x_i\|_p$). The idea is that we will assume the label for $x_i$ and its nearest neighbor $x_i^{(1)}$ are both reasonably accurate, and thus can serve as an estimate of the threshold for which no penalty should occur.

$$\left| f^*(x_i) - f^*(x_i^{(1)}) \right| \approx \left| y_i - y_i^{(1)} \right| \tag{4}$$

Since we are using the nearest neighbors to define a threshold for which no changes should occur, we must also factor in the density of points in each region. A large radius in high density areas would under-regularize the model if the radius extended out past some $\beta \cdot \|x_i - x_i^{(1)}\|_p$ where $\beta \in \mathbb{R}_{>0}$ scales the radius.

To deal with this, we will allow each data point $x_i$ to have a different radius $\Delta_{x_i} < \epsilon_i$ that is based on its nearest neighbor. For convenience we will denote $\Delta_{y_i} = f_\theta(x) - f_\theta(x+\Delta_x)$, and our regularization term will be given a shorthand $\Omega$, as defined in Equation 5.

$$\mathop{\mathbb{E}}_{\Delta_x : \|\Delta_x\|_p < \beta \cdot \|x_i - x_i^{(1)}\|_p} \left[ \ell \left( \Delta_{y_i} \cdot \mathbb{1} \left[ |\Delta_{y_i}| > \left| y_i - y_i^{(1)} \right| \right] \right) \right] \tag{5}$$

The value of $\Omega$ is now adaptive to the local density of the input space by the nearest neighbor defined radius, and imposes no penalty when the perturbed difference $|\Delta_{y_i}|$ is smaller than the difference in response between a point $x_i$ and its nearest neighbor. This captures our intuition that some minimal amount of variation in response is to be expected, and we only wish to penalize unreasonably large deviations. The final robust loss we would minimize is then $\ell_R(y, x) = \ell(y - f_\theta(x)) + \lambda \cdot \Omega$.

### 3.2 An Optimization Connection

We have now walked through our process of deriving the Adversarial Numerical Stability Regularization (ANSR) term in Equation 5. Given our derivation, we take a moment to address some insights and intuition into its nature from an optimization perspective.

The primary goal of our ANSR term is to incorporate a prior belief that some minimum degree of flexibility is necessary for the model to accurately learn the underlying function, and should only be penalized for response changes that are larger than necessary.

A way of phrasing this belief from an optimization perspective is that we want our learned function $f_\theta$ to be Lipschitz continuous with respect to some constant $K$. This would mean for any input $x$ and perturbation $\Delta$, it would hold that $|f_\theta(x) - f_\theta(x+\Delta)| < K \cdot \|x - (x+\Delta)\|_p = K \cdot \|\Delta\|_p$. The Lipschitz property directly captures that the change in outputs should be no more than a factor of $K$ times the change in inputs. It immediately follows that if a $f_\theta$ is $K$-Lipschitz, then an adversarial attack's success is bounded by $\leq K \cdot \|\Delta\|_p$. This property is normally desired in order to make stronger statements about the convergence rate of an optimization procedure or to better select a step-size during gradient decent. Instead we are using Lipschitz continuity as a goal for the function learned, rather than an assumption of the function being optimized.

In greater detail, our goal is really that $f_\theta$ be *locally* Lipschitz continuous, such that $\forall x \in \mathbb{R}^d$ there exists an associated $K_x$ such that $|f_\theta(x) - f_\theta(x+\Delta)| < K_x \cdot \|\Delta\|_p$. This is necessary because the true rate of change for the underlying function will vary throughout the function space. Thus the global Lipschitz constraint would only protect the area of the function that has the maximal rate of change, and fail to bound the safety of flatter and smoother regions of the function space. Equation 4 serves the same purpose as $K_x$ (a local measure of the amount of allowed change), and the $\beta \|x_i - x_i^{(1)}\|_p$ term in the expectation estimates the neighborhood and magnitude of $\Delta$. Locally Lipschitz is normally a weaker assumption (globally Lipschitz implies locally Lipschitz, but not vice versa), but makes our optimization problem more difficult. We must estimate an additional set of parameters embodied by $K_x$ for all points in the space, which limits the degree which we can accurately estimate them. This can be particularly difficult for points in the tail of the distribution. Further, we are not aware of a method to enforce the solution to be locally Lipschitz. While a method was recently proposed for enforcing global $K$-Lipschitz (Gouk et al. 2018), it is not applicable for the reasons just discussed. Devising such techniques is an open avenue for future work which may further improve our ANSR technique.

## 4 Methodology

Following Biggio, Fumera, and Roli (2014), we will define the threat-model by which our hypothetical adversary acts for our evaluation. A white-box scenario will be used, where we assume our adversary knows our model $f_\theta(\cdot)$, its weights $\theta$, and the training data. The adversary's goal will be to perturb an input $x_i$ into $\tilde{x}_i$ such that the squared error between $y_i$ and $f_\theta(\tilde{x}_i)$ is maximized[2]. The adversary's perturbation will be limited to the $L_\infty$ ball of radius $\epsilon = 0.1$ (i.e., $\|x_i - \tilde{x}_i\|_\infty < \epsilon$), but the adversary may otherwise alter any feature as they see fit. Implicit in most prior works is the assumption that the defender knows the value of $\epsilon$, and this assumption is an integral part of the Gradient Regularization defense (Lyu, Huang, and Liang 2015; Simon-Gabriel et al. 2018; Yu et al. 2018). For parity with prior works, we will continue with this assumption.

---

[2]Implicit in this assumption is that the perturbation has not meaningfully changed the true response $y_i$. This is not absolutely true, but since we do not know the generating function, is assumed for simplicity.

Having defined the threat model under which we will operate, we will now enumerate the data sets, attacks, hyperparameter optimization, and other minutia of our methodology.

## 4.1 Data

Experiments were run on the following regression data sets. Prior to use, each data set was split into training, validation, and test sets. The independent variables for each data set were then normalized using training set statistics. All data sets were obtained from the UCI Machine Learning repository, and a summary of their properties is given in Table 1.

Table 1: Size, dimensionality, and sparsity of data sets.

| Data Set | N | D | Sparsity |
|---|---|---|---|
| Boston Housing | 506 | 13 | 12.82% |
| Communities and Crime | 2215 | 101 | 3.23% |
| CT Slices | 53 500 | 384 | 64.95% |
| Malware | 107 856 | 482 | 88.63% |
| Condition Based Maintenance | 11 934 | 16 | 0.40% |

**Boston Housing** In the Boston house price data of Harrison and Rubinfeld (1978), the target variable is the median value in thousands of dollars of owner-occupied homes in the area of Boston, Massachusetts.

**Communities and Crime** The second data set we evaluate our approach on is the Communities and Crime Unnormalized data set (Redmond and Highley 2010). The number of murders in 1995 is the target variable, and variables include potential factors such as percent of housing occupied, per capita income, and police operating budget. Independent variables from the original data set that contained missing values were dropped.

**Relative Location of CT Slices on Axial Axis** The third data set we evaluate our defense on is the Relative Location of CT Slices on Axial Axis data set (Graf et al. 2011). The data consists of a set of 53500 CT images from 74 different patients where each CT slice is described by two histograms in polar space. The histograms describe the location of bone structures in the image and the location of air inclusions inside of the body. The independent variables consist of the information contained in the two histograms, and the target variable is the relative location of an image on the axial axis.

**Malware** The fourth data set we evaluate on is the Dynamic Features of VirusShare Executables data set from Huynh, Ng, and Ariyapala (2017) which contains the dynamic features of executables collected by VirusShare between November 2010 and July 2014. The target variable is a risk score between 0 and 1.

**Condition Based Maintenance** The Condition Based Maintenance of Naval Propulsion Plants data set consists of results from a numerical simulator of a naval vessel characterized by a gas turbine propulsion plant (Coraddu et al. 2016). This data set has two target variables, the gas turbine's compressor decay state coefficient and the gas turbine's turbine decay state coefficient. As such we will treat this as two different regression data sets that use the same feature set.

## 4.2 Attacks

**Fast Gradient Sign Method** The Fast Gradient Sign Method (FGSM) is a $L_\infty$ bounded attack (I. J. Goodfellow, Shlens, and Szegedy 2015; Kurakin, I. Goodfellow, and S. Bengio 2017). Given a training data point $(x, y)$, cost function $J$, and model parameters $\theta$, FGSM computes an adversarial example $\tilde{x}$ as

$$\tilde{x} = x + \epsilon \text{sign}(\nabla_x J(\theta, x, y))$$

where $\epsilon$ is a parameter controlling attack magnitude. FGSM is included as a simple one-step attack baseline.

**Projected Gradient Descent** The Projected Gradient Descent (PGD) attack is a more powerful multi-step variant of FGSM (Kurakin, I. Goodfellow, and S. Bengio 2017). A $Q$-step PGD attack is defined recursively as

$$x^0 = x$$
$$x^{q+1} = \text{clip}_{x,\rho} \left( x^q + \epsilon \text{sign}(\nabla_x J(\theta, x, y)) \right)$$
$$\tilde{x} = x^Q$$

where $\text{clip}_{x,\rho}(x^a)$ is the element-wise clipping of $x^a$ to $[x - \rho, x + \rho]$. Because PGD has been argued as an effective and near-optimal first-order attack method (Madry et al. 2018), we use it as our stronger baseline for an iterative attack.

**Attack Settings** In our experiments, we evaluate performance against both an FGSM attack with magnitude parameter $\epsilon = 0.1$ and a 10-step PGD attack with step magnitude parameter $\epsilon = 0.025$ and clipping radius $\rho = 0.1$. Setting the FGSM magnitude parameter equal to the PGD clipping radius ensures that the attack radius is the same across both FGSM and PGD.

## 4.3 Baselines

Our baseline model will be one trained using the Mean Squared Error (MSE) loss function $\ell(a) = a^2$. Without any defenses implemented, this would correspond to the normal case when a person builds a model without any adversarial assumptions. For each potential defense, we will compare its MSE when under attack to the baseline, as well as its accuracy when no attack has occurred. Our primary comparison point will be our new defense ANSR, but we will also include two other baseline defenses to compare with.

One possible alternative hypothesis to our ANSR approach is that our use of the indicator function in (5) will allow too much variation and instability in larger areas of the search space. That instead of penalizing numerical instability, we will instead form some kind of approximate robust-estimator, where the loss surface changes more significantly as the absolute difference $a$ increases (Huber 1981). Prior work has proposed that the standard softmax loss is a potential enabler of adversarial examples in the classification scenario and has combined a change in classification loss with a non-negativity constraint (Fleshman et al. 2018) to increase adversarial robustness. Since robust-estimators obtain better fits when outliers exist in the training distribution, it would make sense that they could improve defenses against adversarial attacks. By avoiding the impact of outliers, they also avoid learning more complex functions (with more room for adversaries to operate). As far as we are aware, our work is the first to directly test if a simple change in loss function can reduce adversarial attack success.

To test this hypothesis, we will train a model that replaces the MSE loss with the Pseudo-Huber loss (6), which is a robust-estimator.

$$\ell_\delta(a) = \delta^2 \left( \sqrt{1 + (a/\delta)^2} - 1 \right) \quad (6)$$

The Pseudo-Huber loss is a smooth approximation of the Huber loss function, where $\delta$ is a loss steepness parameter (Barron 2017). If ANSR is nothing more than a crude approximation of a robust estimator, we would expect the Pseudo-Huber to perform better than ANSR, and that ANSR and Pseudo-Huber would be highly correlated in their results.

As discussed in section 2, Gradient Regularization is a common approach to improving adversarial robustness, which has been re-invented multiple times and can be interpreted as penalizing the absolute condition number of the network. If ANSR does work by improving numerical stability as intended, but our hypothesis that a minimum degree of unpenalized flexibility should exist is wrong, we would expect to see Gradient Regularization perform better. When regularizing with respect to attacks measured by the $L_\infty$ norm, the regularization of the gradient becomes (7).

$$\ell_\sigma(a) = \ell(a) + \sigma \left\| \frac{\partial \ell(a)}{\partial x} \right\|_1 \quad (7)$$

We use the notation of Lyu, Huang, and Liang (2015) where $\sigma$ is a regularization strength parameter. Under the frameworks where Gradient Regularization has been proposed, $\sigma$ would be set to the maximum perturbation distance $\epsilon$.

Finally, we note that all three of the primary defenses we will evaluate, our ANSR, Pseudo-Huber, and Gradient Regularization, can be merged into a combined defense. The primary loss can be Pseudo-Huber, combined with (7) and (5) as regularization terms.

### 4.4 Hyperparameter Tuning

Hyperparameters for the defense as well as for baselines were selected by cross-validation using randomized search (Bergstra and Y. Bengio 2012) and mean squared error (MSE) as an evaluation metric. For the Pseudo-Huber baseline, loss steepness $\delta$ was uniformly sampled from the interval $[0.01, 16.0]$. Normally for the Gradient Regularization defense baseline, since we have defined the attack radius as $0.1$, we would set $\sigma = 0.1$. This is because the defense is derived with $\sigma$ being equal to the attack radius in the initial derivation. To make the Gradient Regularization defense an even stronger baseline, we sample uniformly the regularization strength $\sigma$ from the interval $[0.01, 16.0]$. This allows Gradient Regularization to be potentially stronger than necessary and maximally quash adversarial attack strength. For our ANSR defense, the radius scaling parameter $\beta$ was uniformly sampled from the interval $[0.5, 8.0]$ and the stabilization strength $\lambda$ was uniformly sampled from the interval $[0.1, 10.0]$. The expectation in equation (5) is estimated using 100 samples.

### 4.5 Network Architecture and Training

The network architectures in all experiments consist of an input layer, a single hidden layer of size equal to the input layer size, and a single node output layer. Rectified linear units were used as hidden layer activation functions, and Adam was used as the optimization algorithm (Kingma and Ba 2014). For the Malware data set, the output was constrained to $[0, 1]$ using a sigmoid activation on the output layer.

## 5 Results and Discussion

We will now compare the results of our ANSR defense against others on the specified datasets, where we will see it has the best overall performance. Then we will investigate properties of the attacks and defenses, and pursue alternative hypotheses to our numerical stability theory.

### 5.1 Optimizing Against Adversary

We begin examining ANSR and other approaches when attacked with a maximum perturbation radius of $\epsilon = 0.1$ under the $L_\infty$ distance. The Mean Squared Error on the test set for all six data sets under evaluation are shown in Table 2. Under both the FGSM and PGD attacks, we can see that our ANSR defense performs best on five out of six corpora. Across all data sets we see that FGSM results have the same behavior as those generated with PGD, with moderately lower MSE. As such we will focus our discussion around PGD.

Table 2: The mean and standard deviation (in parentheses) of the Mean Squared Error on the test set for each data set. The means and standard deviations were estimated by randomly initializing and retraining each network 6 times. The right most columns show the results for each defense (and no-defense) under no attack, FGSM, and PGD based attacks. Best results shown in **bold** for each column, with models optimized to reduce error under PGD.

| Data Set | Defense | No Attack | FGSM | PGD |
|---|---|---|---|---|
| Boston | ANSR | 1.48E+01 (7.19E-01) | **2.48E+01 (1.20E+00)** | **2.50E+01 (1.17E+00)** |
| | Pseudo-Huber | 1.44E+01 (6.63E-01) | 4.31E+01 (1.85E+00) | 4.36E+01 (1.76E+00) |
| | Gradient Regularization | **1.32E+01 (6.19E-01)** | 3.69E+01 (1.91E+00) | 3.73E+01 (2.00E+00) |
| | No Defense | 1.47E+01 (1.05E+00) | 4.83E+01 (1.91E+00) | 4.89E+01 (1.97E+00) |
| Crime | ANSR | **2.52E+02 (3.79E+01)** | **2.86E+02 (4.13E+00)** | **2.87E+02 (4.18E+00)** |
| | Pseudo-Huber | 2.95E+02 (3.68E+01) | 3.53E+02 (3.75E+01) | 3.54E+02 (3.75E+01) |
| | Gradient Regularization | 1.76E+03 (1.82E+02) | 1.91E+03 (1.84E+02) | 1.91E+03 (1.84E+02) |
| | No Defense | 1.84E+03 (1.82E+02) | 2.01E+03 (1.81E+02) | 2.01E+03 (1.81E+02) |
| CT Slices | ANSR | 1.45E+02 (4.42E+01) | **3.17E+02 (4.12E+01)** | **3.21E+02 (4.15E+01)** |
| | Pseudo-Huber | 9.64E+01 (8.71E+00) | 5.39E+02 (6.70E+01) | 5.45E+02 (6.80E+01) |
| | Gradient Regularization | 1.02E+02 (1.59E+01) | 3.71E+02 (4.23E+01) | 3.76E+02 (4.22E+01) |
| | No Defense | **9.41E+01 (6.00E+00)** | 5.16E+02 (2.41E+01) | 5.22E+02 (2.46E+01) |
| Malware | ANSR | 6.24E-02 (4.62E-03) | 2.17E-01 (2.68E-02) | 2.51E-01 (4.17E-02) |
| | Pseudo-Huber | 6.62E-02 (6.57E-03) | 2.55E-01 (3.69E-02) | 2.87E-01 (4.29E-02) |
| | Gradient Regularization | 7.21E-02 (1.17E-02) | **1.69E-01 (4.37E-02)** | **1.57E-01 (3.39E-02)** |
| | No Defense | **6.70E-02 (7.10E-03)** | 2.78E-01 (4.31E-02) | 3.22E-01 (5.31E-02) |
| CBM Comp | ANSR | 3.07E-04 (6.91E-05) | **1.16E-02 (2.57E-03)** | **1.18E-02 (2.50E-03)** |
| | Pseudo-Huber | 3.51E-04 (6.54E-05) | 1.42E-02 (3.04E-03) | 1.44E-02 (3.01E-03) |
| | Gradient Regularization | 3.57E-04 (4.54E-05) | 1.41E-02 (2.67E-03) | 1.42E-02 (2.69E-03) |
| | No Defense | **3.38E-04 (2.06E-05)** | 1.72E-02 (4.83E-03) | 1.73E-02 (4.83E-03) |
| CBM Turb | ANSR | 1.62E-04 (1.17E-04) | **1.26E-02 (4.54E-03)** | **1.27E-02 (4.49E-03)** |
| | Pseudo-Huber | **1.26E-04 (4.13E-05)** | 1.53E-02 (1.84E-03) | 1.54E-02 (1.82E-03) |
| | Gradient Regularization | 1.52E-04 (9.34E-05) | 1.32E-02 (3.10E-03) | 1.33E-02 (3.09E-03) |
| | No Defense | 1.39E-04 (8.87E-05) | 1.32E-02 (2.66E-03) | 1.32E-02 (2.68E-03) |

These results allow us to infer conclusions regarding a number of potential hypotheses regarding performance and the method by which ANSR provides improved robustness to adversarial attacks.

First, we can see that the Pseudo-Huber loss improves the model's resistance to adversarial attack in four out of six datasets. This tells us that the robust regression hypothesis has merit, and shows that we can often improve results by giving greater deference to the choice in target loss to optimize for. The Pseudo-Huber's improvement was most significant on the Crime data set, which has a target response $y$ in the range of $[0, 1000]$. This is the largest response range of our corpus, and Pseudo-Huber produces results close to that of our ANSR defense while also providing a $6.2\times$ reduction to MSE when no attack is present. In this case, ANSR also provides a better reduction in un-attacked MSE by a factor of 7.3. This would seem to confirm that outliers can have a significant impact on a model's susceptibility to adversarial examples.

Pseudo-Huber loss only produces an improvement of $1.1\times$ to MSE under PGD attack on the Boston data set while performing worse than no defense on the CT Slices dataset, while ANSR has more dramatic $2.0\times$ and $1.6\times$ improvements for each respective corpus. In addition, ANSR and Pseudo-Huber have similar performance on Crime and Malware corpora, but dissimilar performance on the Boston and CT Slices data sets. As such, we do not believe ANSR's benefit is just an artifact of robust regression, and thus ANSR is more than a robust regression technique.

The second alternative hypothesis was that ANSR may be a less effective approach of penalizing numerical instability than Gradient Regularization's minimization of the absolute condition number of the model. However, we see on the Crime data set that Gradient Regularization has $1.05\times$ improvement in MSE, where ANSR has a dramatically improved $7.0\times$. Gradient Regularization is also less improved on the Boston data set and still trails on the CT Slices data set. These results would appear to support our prior belief that in the regressions setting, a minimum degree of model flexibility should be allowed without penalization.

The only case where ANSR does not provide the best MSE under attack by PGD is on the Malware dataset. In this case, Gradient Regularization performs best with a $2.1\times$ reduction, followed by ANSR at a factor 1.3 and Pseudo-Huber at 1.1. Gradient Regularization's success in this case is because we have allowed it to over-regularize by choosing values of $\sigma > \epsilon = 0.1$. In this case, cross-validation found $\sigma = 2.9$ minimized the PGD attack MSE, a value $29\times$ greater than what standard Gradient Regularization would have selected.

ANSR appears to be the best individual regularization approach against adversarial attacks across the majority of datasets. It uniformly dominates the Pseudo-Huber loss under attack, and with one exception does the same compared to an enhanced Gradient Regularization that is allowed to over-select the strength of $\sigma$. Though not a design goal, ANSR even improves non-attacked MSE on three out of six datasets, by up to a factor of $7.3\times$.

Because the three defenses we are evaluating target different aspects of the adversarial problem, we can further combine them to produce a combined defense. The results are shown in Table 3, where we see the combined defense improves the results on the CT Slices and CBM Turbine datasets — while still being competitive on all other data sets. The improvement gives us further confidence that ANSR tackles different aspects than just robust regression or the prior Gradient Regularization approach.

Table 3: Results when combining all three defenses into a single combined defense. The MSE means and standard deviations (in parentheses) were estimated by randomly initializing and retraining each network 6 times. Results in **bold** when better than than any individual defense from Table 2.

| Data Set | Attack | | |
| --- | --- | --- | --- |
| | No Attack | FGSM | PGD |
| Boston | 1.63E+01 (9.97E-01) | 2.50E+01 (1.38E+00) | 2.52E+01 (1.36E+00) |
| Crime | 3.15E+02 (2.50E+01) | 3.47E+02 (2.57E+01) | 3.48E+02 (2.57E+01) |
| CT Slices | 1.30E+02 (3.59E+01) | **2.66E+02 (3.59E+01)** | **2.70E+02 (3.60E+01)** |
| Malware | 1.81E-01 (1.29E-01) | 1.93E-01 (1.13E-01) | 1.94E-01 (1.11E-01) |
| CBM Comp | 3.21E-04 (8.16E-05) | 1.21E-02 (3.46E-03) | 1.22E-02 (3.51E-03) |
| CBM Turb | 1.98E-04 (1.33E-04) | **1.17E-02 (3.25E-03)** | **1.19E-02 (3.18E-03)** |

### 5.2 Properties of Adversarial Perturbations

Now that we have shown that ANSR provides improved defense against adversarial attacks on regression problems, we seek to further elucidate properties and aspects of its effectiveness. We first begin by looking at the distribution of perturbation magnitudes. Since we are evaluating adversarial regression, errors are a continuous function of the output — rather than a discrete change once one class probability surpasses another like the classification setting. As such, reductions in MSE could occur in multiple ways. The two general cases are shifting the entire distribution of perturbations (i.e., reducing the maximum successful attack magnitude), and right-skewing of the distribution (i.e., changing the frequency of success types). A defense will ideally exhibit both of these properties on a consistent basis.

The distributions of PGD magnitudes are shown in Figure 1 for each data set. In this figure we can see that when no defense is present, the perturbation magnitude has no particular distribution. For Crime it is unimodal with mild right skew, on Boston and Malware it is bimodal and almost symmetric, more erratic with left-skew in distribution on the CT Slices corpus, and simply bimodal on the CBM data sets.

All three defenses make some improvement in shifting the distribution of perturbation magnitude to the left (i.e., smaller), but not always. On Crime, Gradient Regularization only increases the right-skew, and on Malware, Pseudo-Huber has the same effect.

An overall trend is that the Pseudo-Huber loss and Gradient Regularization have relatively little impact on the nature of the distribution (i.e., relatively similar shapes). By comparison, ANSR changes every distribution but CBM Turbine to be right skewed and reduces the maximum attack magnitude. In all but the Malware corpus, this includes moving the majority of the distribution to a near-zero perturbation magnitude. ANSR is the only method to have such significant impact on the distribution of perturbations, and we believe this is a result of directly tackling the numerical stability hypothesis.

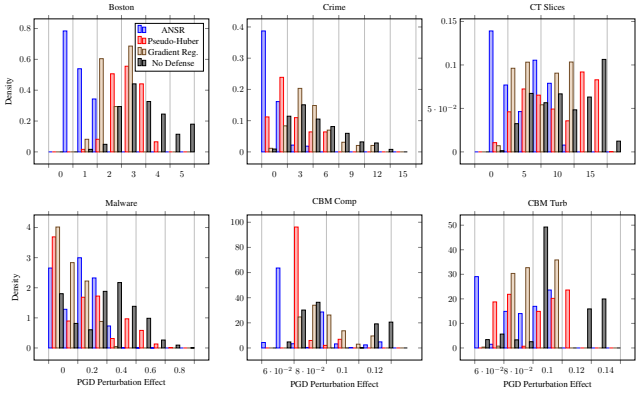

Figure 1: Magnitude of PGD's perturbation on the response of the model when each defense is used in training. The y-axis shows the density of the distribution, and the x-axis the binned magnitude of the perturbation. The farther left the distribution is, the better it has performed.

Given these results, we also wish to investigate two alternative hypotheses to the source of adversarial perturbations in the regression case: a relationship between perturbation magnitude (i.e., $|\hat{y}_{\text{PGD}} - y|$) and target response, and lower density regions of the data. If the source was with relation to response magnitude, we could attribute improvements to a smoothing of the loss surface by ANSR and Gradient Regularization (Yu et al. 2018). If connected to point density, we could attribute improvement to ANSR connecting information between points and their density neighborhood. If neither strongly holds, we believe this leaves the numerical stability and optimization theory discussed in subsection 3.2 as the best explanation for ANSR's improvement.

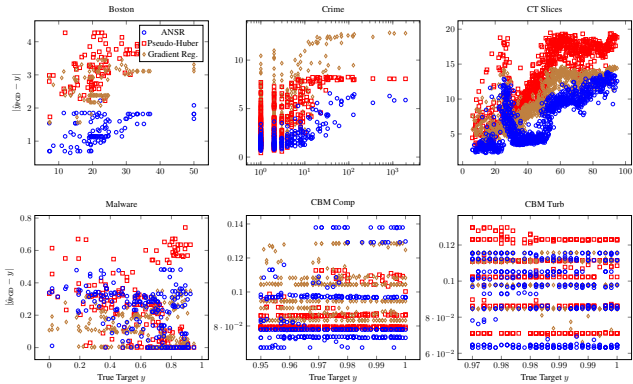

Figure 2: Plots of the magnitude of the perturbation under PGD attack (y-axis) against the true target value $y$ (x-axis).

The first hypothesis, that as $y$ becomes larger the relative change in the size of the perturbation under PGD stays the same or increases, is investigated in Figure 2. The results provide some evidence of a consistent trend that larger $y$ values produce larger perturbations. This however does not capture the whole story. On the CT Slices data set we can see that there is some distinct property of the datums at $y \approx 27$ that causes a large spike in susceptibility to attack across all defenses. On Malware we see ANSR and Pseudo-Huber have both positive and negative correlation with the response $y$, and on Crime we see both these methods have no impact on the perturbation magnitude for $y$ over two orders of magnitude in range. Overall, based on the variety of results we do not believe the response range $y$ is a causative source of successful adversarial attacks.

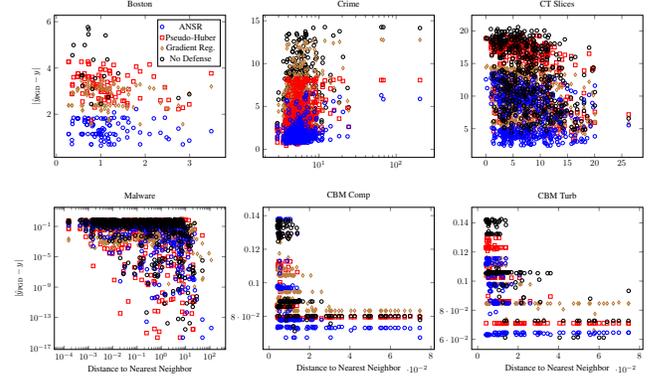

Figure 3: Plots of the magnitude of the perturbation under PGD attack (y-axis) against each point's distance to its nearest neighbor (x-axis).

The second hypothesis is that in highly dense regions of the input space, the function is well modeled and leaves little room for successful attacks, and that success lies in the lower density regions of the space. In Figure 3 we plot the perturbation magnitude as a function of nearest neighbor distance. Across all three defensive techniques, we see no particular relationship between the magnitude of PGD attack success and the density of the attacked point. On the CBM data sets we even have evidence of the opposite of the hypothesis. In both cases the most successful adversarial attacks exist only in the higher density points.

Given these results, the numerical stability hypothesis appears to be the best current explanation for ANSR's success. It is the only defense to meaningfully change the distribution of attack successes, and our investigation of two alternative hypotheses fails to provide contradictory evidence.

## 6 Conclusion

In this work we have introduced ANSR, a new regularization approach to combat adversarial attacks for regression applications that out-performs other approaches. It is inspired by encouraging numerical stability with theoretical connections to locally Lipschitz continuity. Alternative hypotheses to ANSR's success did not displace our numerical stability theory.

## Acknowledgements

Special thanks to Drew Farris for his support of this work and to Arash Rahnama for interesting conversations and insights.